\documentclass[10pt,twocolumn,letterpaper]{article}

\usepackage[pagenumbers]{cvpr} 

\definecolor{cvprblue}{rgb}{0.21,0.49,0.74}
\usepackage[pagebackref,breaklinks,colorlinks,allcolors=cvprblue]{hyperref}
\usepackage{svg}
\usepackage{array}
\usepackage{float}
\usepackage{arydshln}
\usepackage{amsmath}
\usepackage{algorithm}
\usepackage{algpseudocode}
\usepackage{multirow}
\usepackage{makecell}
\usepackage{xcolor}
\usepackage{booktabs} 

\def\paperID{15900} 
\def\confName{CVPR}
\def\confYear{2025}

\title{NitroFusion: High-Fidelity Single-Step Diffusion\\through Dynamic Adversarial Training}

\author{%
  Dar-Yen Chen$^{1, 2}$\:\; Hmrishav Bandyopadhyay$^{1}$\:\; Kai Zou$^{2}$\:\; Yi-Zhe Song$^{1}$\\[0.2cm]
  $^{1}$SketchX, CVSSP, University of Surrey \quad
  $^{2}$NetMind.AI \\
  \small{\texttt{\{d.chen, h.bandyopadhyay, y.song\}@surrey.ac.uk    }}    
  \small{\texttt{kz@netmind.ai}}\\
  \small{\url{https://chendaryen.github.io/NitroFusion.github.io}}
}

\begin{document}
\twocolumn[{
\renewcommand\twocolumn[1][]{#1}%
\maketitle

\thispagestyle{empty}
\vspace{-1.1cm}
\begin{center}
    \centering
    \captionsetup{type=figure}
    \includegraphics[width=1\linewidth]{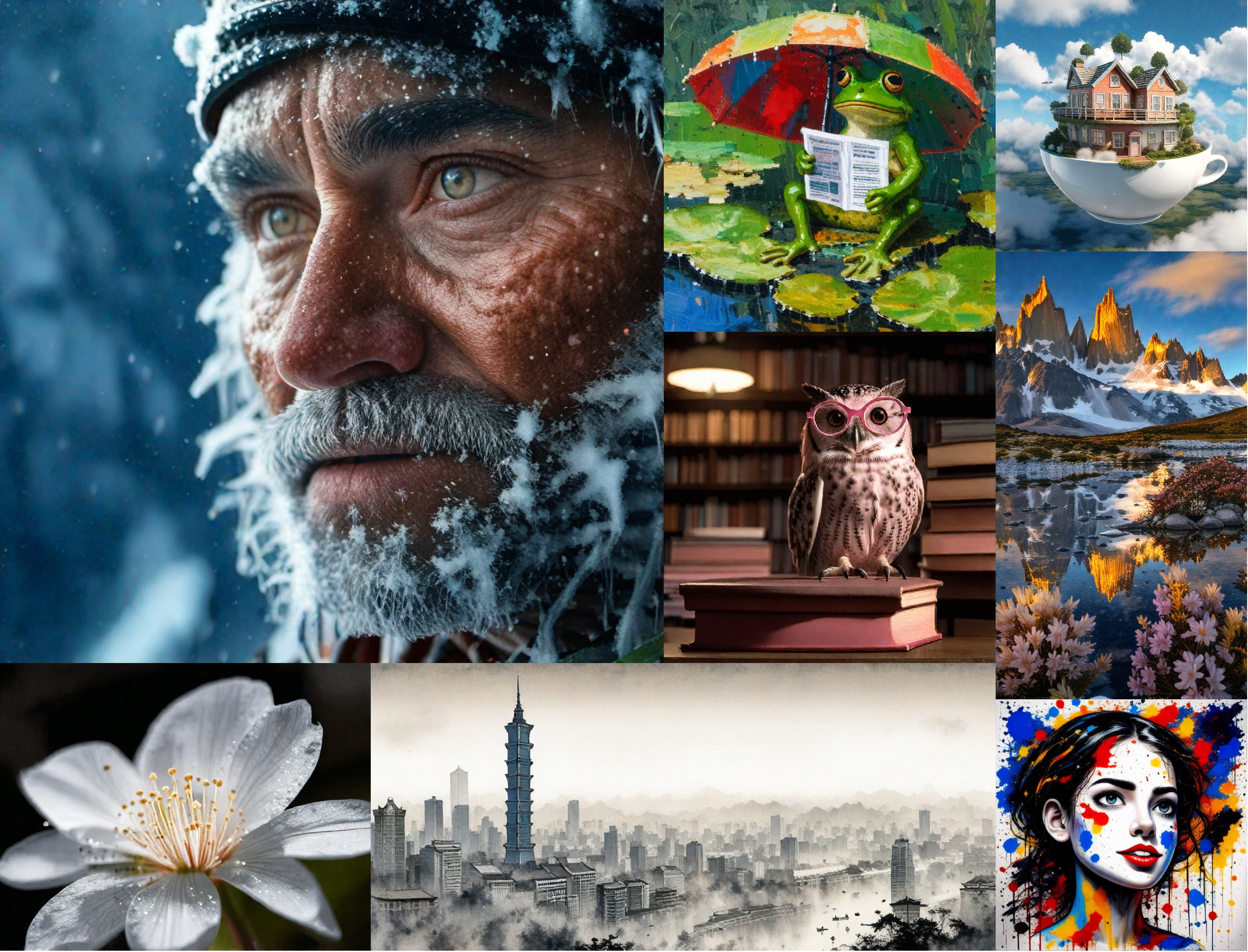}
    \vspace{-7mm}
    \captionof{figure}{
    Our one-step diffusion pipeline generates vibrant and photorealistic images with exceptional detail in a single inference step, broadening the potential for text-to-image synthesis in applications like real-time interactive systems. 
    }
    \vspace{-0.07cm}
    \label{fig:banner}
\end{center}
}]
\begin{abstract}

We introduce NitroFusion, a fundamentally different approach to single-step diffusion that achieves high-quality generation through a dynamic adversarial framework. While one-step methods offer dramatic speed advantages, they typically suffer from quality degradation compared to their multi-step counterparts. Just as a panel of art critics provides comprehensive feedback by specializing in different aspects like composition, color, and technique, our approach maintains a large pool of specialized discriminator heads that collectively guide the generation process. Each discriminator group develops expertise in specific quality aspects at different noise levels, providing diverse feedback that enables high-fidelity one-step generation. Our framework combines: (i) a dynamic discriminator pool with specialized discriminator groups to improve generation quality, (ii) strategic refresh mechanisms to prevent discriminator overfitting, and (iii) global-local discriminator heads for multi-scale quality assessment, and unconditional/conditional training for balanced generation. Additionally, our framework uniquely supports flexible deployment through bottom-up refinement, allowing users to dynamically choose between 1-4 denoising steps with the same model for direct quality-speed trade-offs. Through comprehensive experiments, we demonstrate that NitroFusion significantly outperforms existing single-step methods across multiple evaluation metrics, particularly excelling in preserving fine details and global consistency.

\end{abstract}    
\vspace{-2mm}
\section{Introduction}
\label{sec:intro}

Recent advances in accelerated diffusion models~\cite{luhman2021knowledge, gu2023boot, meng2023distillation, kim2023consistency, heek2024multistep, yan2024perflow, xu2024accelerating, zhou2024score} have demonstrated that high-quality image generation is possible with dramatically reduced step counts. While several approaches now achieve one-step generation~\cite{sauer2023adversarial, zhang2023hipa, nguyen2023swiftbrush, lin2024sdxl, ren2024hyper, yin2024onestep, yin2024improved}, they face significant challenges in matching the quality of multi-step methods, particularly in preserving fine details and ensuring global coherence. This quality gap has limited the practical adoption of single-step methods, especially in applications requiring both speed and high fidelity.

The core challenge in single-step diffusion lies in compressing an entire denoising trajectory~\cite{zheng2024trajectory, liu2022flow} into a single transformation. Traditional approaches based on distillation~\cite{salimans2022progressive, song2023consistency} struggle because they attempt to directly match intermediate states or distributions, leading to blurry outputs and loss of detail. Recent adversarial methods~\cite{Goodfellow2014GAN, sauer2023adversarial, sauer2024fast, yin2024improved} show promise but face training instability and diversity collapse when pushed to single-step generation.

NitroFusion introduces a fundamentally different approach to single-step diffusion through a dynamic adversarial framework. Consider how a panel of art critics evaluates a painting -- each critic specializes in different aspects like composition, color, technique, and detail. Similarly, rather than relying on a single discriminator that can quickly become overconfident~\cite{durugkar2017generative, nguyen2017dual, neyshabur2017stabilizing, choi2022mclgan}, we maintain a large, dynamic pool of specialized discriminator groups that operate on top of a frozen UNet backbone~\cite{rombach2022high}. Just as a diverse panel of critics provides more comprehensive feedback than a single judge, our ensemble of discriminators guides the generator toward high-quality outputs by providing specialized feedback at different noise levels~\cite{lin2024sdxl} and spatial scales.

Our framework implements this insight through three technical innovations: 
(i)~a dynamic discriminator pool architecture where we leverage the teacher model's UNet encoder as a frozen feature extractor, with multiple lightweight discriminator groups $\mathcal{H}_{t^*}$ specialized for different noise levels $t^*$ to improve generation quality, 
(ii)~a strategic refresh mechanism that randomly re-initializes $\sim$1\% of discriminator heads while preserving the collective knowledge distribution across the pool to prevent discriminator overfitting -- a common failure mode in GAN training -- while maintaining stable adversarial feedback, and 
(iii)~a multi-scale strategy with dual training objectives where global heads and local heads are compartmentalized in a 1:2 ratio, with global heads assessing overall image coherence at resolution $H{\times}W$ and local heads examining fine-grained details in patches of size $h{\times}w$. These are further divided as unconditional and prompt-conditional discriminator heads (dual-training) effectively balancing prompt alignment with image coherence.

These technical components work together to solve the fundamental challenges of single-step generation. The dynamic discriminator pool and refresh mechanism work in tandem to maintain a balanced feedback system throughout training -- as established heads provide consistent feedback, the periodic introduction of new heads prevents the system from becoming too rigid or predictable. The multi-scale strategy then complements this dynamic feedback system, enabling our generator to achieve what previous approaches could not: transforming noise into high-quality images in a single step while avoiding the artifacts and quality degradation that typically plague fast generation methods.

Notably, unlike existing approaches~\cite{lin2024sdxl, ren2024hyper, yin2024improved} that require separate models for different step counts, our framework uniquely supports flexible deployment through bottom-up refinement. While we optimize primarily for single-step generation, our model uniquely enables dynamic refinement -- users can simply add steps (up to 4) on-demand if higher quality is desired, all with the same model weights.

Through extensive experimentation, we demonstrate that NitroFusion consistently produces sharper, more detailed images than existing single-step methods. Our approach not only matches but often exceeds the quality metrics of recent fast diffusion models while maintaining the speed advantages of single-step generation. Human evaluation studies further confirm the superior visual quality of our results, particularly in challenging areas like face detail and texture preservation.

Our key contributions include: (i)~a dynamic discriminator pool with specialized discriminator groups to improve generation quality, (ii)~strategic refresh mechanisms to prevent discriminator overfitting, and (iii)~multi-scale strategy with dual training objectives to effectively balance prompt alignment and image coherence. Additionally, we uniquely enable flexible deployment by supporting 1-4 denoising steps with the same model weights.

\vspace{-2mm}
\section{Related Works}

\label{sec:related}
\vspace{-1mm}
\subsection{Timestep Distillation}
\vspace{-1.5mm}
Timestep distillation accelerates inference in diffusion models by reducing the required sampling steps for high-quality output. Standard approaches~\cite{luhman2021knowledge, gu2023boot, zhang2023hipa, meng2023distillation, nguyen2023swiftbrush, heek2024multistep, yan2024perflow, xu2024accelerating, zhou2024score} distil a multi-step teacher model into a student model with fewer steps. A common strategy is to approximate the sampling trajectory, modeled as an ordinary differential equation (ODE), of the teacher model in a reduced step count.
This can be implemented by either preserving~\cite{zheng2024trajectory} the original ODE path at each timestep, or reformulating~\cite{liu2022flow, sauer2023adversarial} and learning a more efficient trajectory directly from the final outputs.
Recent works train a series of such student models that progressively lower sampling steps~\cite{salimans2022progressive, luo2023latent}, while enforcing self-consistency ~\cite{song2023consistency, kim2023consistency}. 
Hyper-SD~\cite{ren2024hyper} further combines ODE-preserving and -reformulating methods. However, these models often face quality degradation due to limited model fitting capacity.
Different from flow-guided distillation, Distribution Matching Distillation (DMD)~\cite{yin2024onestep, yin2024improved} minimizes the Kullback-Leibler (KL) divergence between generated and target distributions to directly match distributions on the sample domain. Despite these advancements, achieving high fidelity in one-step distillation remains challenging, as these models frequently struggle with degradation and instability in extreme low-step settings.

\vspace{-1mm}
\subsection{Adversarial Distillation} 
\vspace{-1.5mm}
Adversarial Diffusion Distillation~\cite{sauer2023adversarial, sauer2024fast} (ADD) incorporates GAN training to address the limitations of MSE-based distillation in the few-step generation, which often leads to blurry outputs. Generally, a pretrained feature extractor~\cite{oquab2023dinov2} is used as the discriminator backbone to obtain stable, discriminative features~\cite{sauer2023icml}. SDXL-Lightning~\cite{lin2024sdxl} for instance, uses the encoder of a pretrained diffusion model as the discriminator backbone, injecting noise prior to the real-vs-fake judgment as a form of augmentation ~\cite{lin2024sdxl}. Recent works ~\cite{kim2023consistency, yin2024improved, dao2024swiftbrushv2makeonestep} further integrate adversarial loss with distillation objectives to improve image fidelity. However, adversarial loss introduces its own challenges, including training instability and reduced diversity~\cite{dao2024swiftbrushv2makeonestep}. Rapid discriminator learning can lead to overconfident assessments, limiting constructive feedback for the generator and causing suboptimal training dynamics. Overcoming these challenges is a primary goal of our work.

\begin{figure*}[htbp!]
\begin{center}
   \includegraphics[trim={5cm 0 7cm 0}, width=0.95\linewidth]{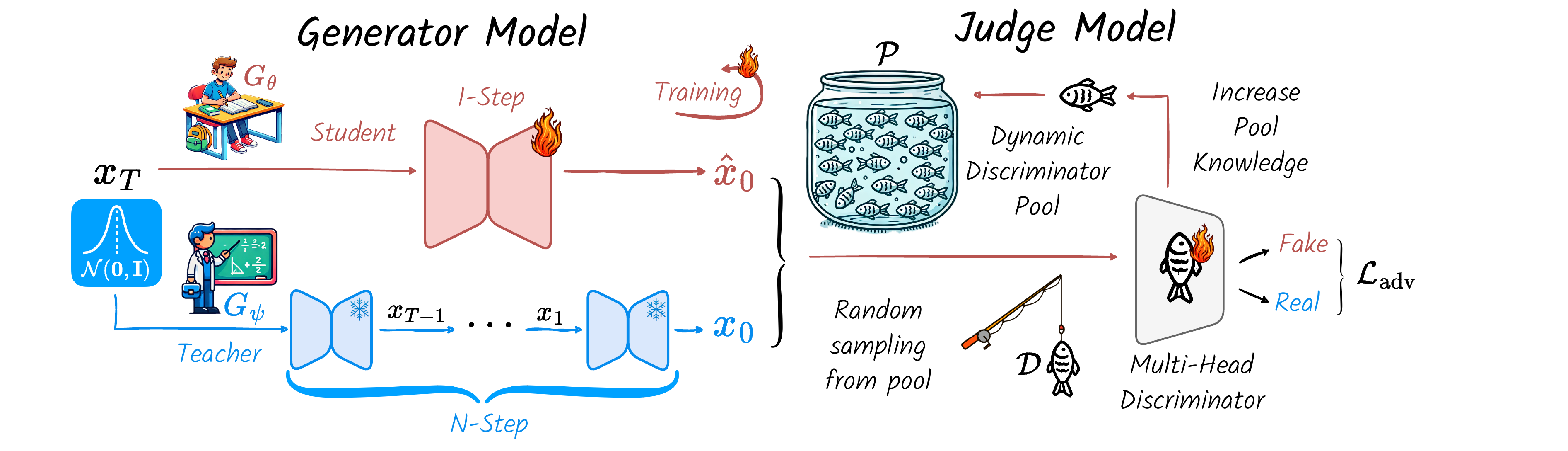}
\end{center}
   \vspace{-7mm}
   \caption{
   Our method distils a multi-step teacher model into an efficient one-step student generator. The Dynamic Adversarial Framework provides dynamic, stable feedback via a large dynamic Discriminator Head Pool, dynamically sampling a subset of heads in each iteration to provide unbiased and stable feedback to judge real or fake, effectively balancing one-step efficiency with high-quality generation.
   }
   \vspace{-4mm}
\label{fig:approach}
\end{figure*}

\vspace{-1mm}
\subsection{Multi-Discriminator Training}
\vspace{-1.5mm}
GANs with multiple discriminators have reduced mode collapse and enhanced training stability through the incorporation of diverse adversarial feedback. Various strategies have been developed to balance multiple discriminator objectives, including \textit{softmax}-weighted ensembles~\cite{durugkar2017generative} and three-player minimax games~\cite{nguyen2017dual}. To address overconfidence in discriminators, Neyshabur \etal~\cite{neyshabur2017stabilizing} applies lower-dimensional random projection for each discriminator, while MCL-GAN~\cite{choi2022mclgan} incorporates multiple choice learning.
StyleGAN-XL~\cite{sauer2022siggraph} and StyleGAN-T~\cite{sauer2023icml} use multiple discriminator heads alongside a frozen, pretrained backbone, enabling feedback across feature pyramids to capture various levels of detail. While these multi-discriminator methods address challenges in GAN training, they remain under-explored in diffusion distillation. Our approach builds upon these insights, introducing a robust adversarial framework to provide diverse and dynamic feedback for high-fidelity one-step diffusion distillation.

\vspace{-1mm}
\section{Methodology}
\label{sec:approach}
\vspace{-1.5mm}

To perform one-step diffusion, we utilize the concept of timestep distillation. In here, a one-step student model is trained to perform at par with a pre-trained multi-step teacher. After training, the one-step student can be used independently for super-fast inference. Unlike conventional methods that rely on score matching~\cite{yin2024onestep} or flow matching~\cite{liu2022flow} to align student and teacher quality, our approach uses adversarial loss only for critiquing teacher and student predictions - akin to a panel of critics that evaluate paintings. This helps us align teacher and student distributions for the student to mimic the teacher in a single step without quality degradation.

Specifically, we propose a Dynamic Adversarial Framework, as: (i) A huge pool of discriminator heads with specialized discriminators for different levels of noise and quality, reducing feedback bias from an otherwise single discriminator set-up. (ii) A periodic pool refresh to randomly re-initialize a sampled set of discriminators to prevent over-fitting, and (iii) multi-scale dual-objective GAN training to reduce artifacts and balance image coherence with prompt alignment. Figures \ref{fig:approach} and \ref{fig:pool_head} illustrate our training pipeline.

\noindent\textbf{Preliminaries}:
Diffusion Models~\cite{ho2020denoising} iteratively refine noise in a data sample by reversing a forward process that progressively transforms an input sample $x_0$ into noise. In this forward process, each noisy sample $x_t$ is obtained from $x_0$ using Gaussian noise $\epsilon \sim \mathcal{N}(0, \mathbf{I})$ at timestep $t \in \{1, \dots, T\}$ as:
\begin{equation}
    x_t = \sqrt{\bar{\alpha}_t} x_0 + \sqrt{1 - \bar{\alpha}_t} \epsilon,
    \label{eq:forward_diffusion}
\end{equation}
where $\bar{\alpha}_t$ is a variance schedule controlling the noise level~\cite{ho2020denoising, song2020score}.
The reverse process, parameterized by a neural network $G_{\theta}$, is trained to predict the noise $\epsilon$ from $x_t$ to reconstruct $x_0$.
Using the predicted noise $\hat{\epsilon} = G_{\theta}(x_t, t)$, $x_0$ is reconstructed as:
\begin{align}
    x_0 = \frac{x_t - \sqrt{1 - \bar{\alpha}_t} \hat{\epsilon}}{\sqrt{\bar{\alpha}_t}}.
    \label{eq:reverse_diffusion_simple}
\end{align}

\vspace{-1mm}
\subsection{One-Step Adversarial Diffusion Distillation}
\vspace{-1.5mm}

Our training pipeline consists of a one-step student (generator) $G_{\theta}$, and a pretrained multi-step teacher model $G_{\psi}$. We initialize the student with pre-trained one-step weights \cite{ren2024hyper, yin2024improved} $\theta$, to reduce the time to converge.
During each training iteration, $G_{\theta}$ and $G_{\psi}$ denoise a noisy sample $x_T \sim \mathcal{N}(0, \mathbf{I})$ to $\hat{x}_0$ and $x_0$ respectively. While this denoising takes multiple steps for the teacher $G_\psi$, our student $G_\theta$ directly denoises $x_T$ to $x_0$ in one step only (see \cref{fig:approach}). The discriminator $\mathcal{D}$ attempts to distinguish $x_0$ as real and $\hat{x}_0$ as fake, constructing the adversarial loss $\mathcal{L}_{\text{adv}}$.  
\begin{equation}
    \mathcal{L}^{G}_\text{adv} = -\mathbb{E}[ \mathcal{D}(\hat{x}_{0})]
\end{equation}
\begin{equation}    
    \mathcal{L}^{D}_\text{adv} = \mathbb{E}[ \mathcal{D}(\hat{x}_{0}) - \mathcal{D}(x_0))]
\end{equation}

\vspace{-1mm}
\subsection{Dynamic Discriminator Pool} 
\vspace{-1.5mm}

\begin{figure}
    \centering
    \includegraphics[trim={1cm 0 0cm 0}, width=\linewidth]{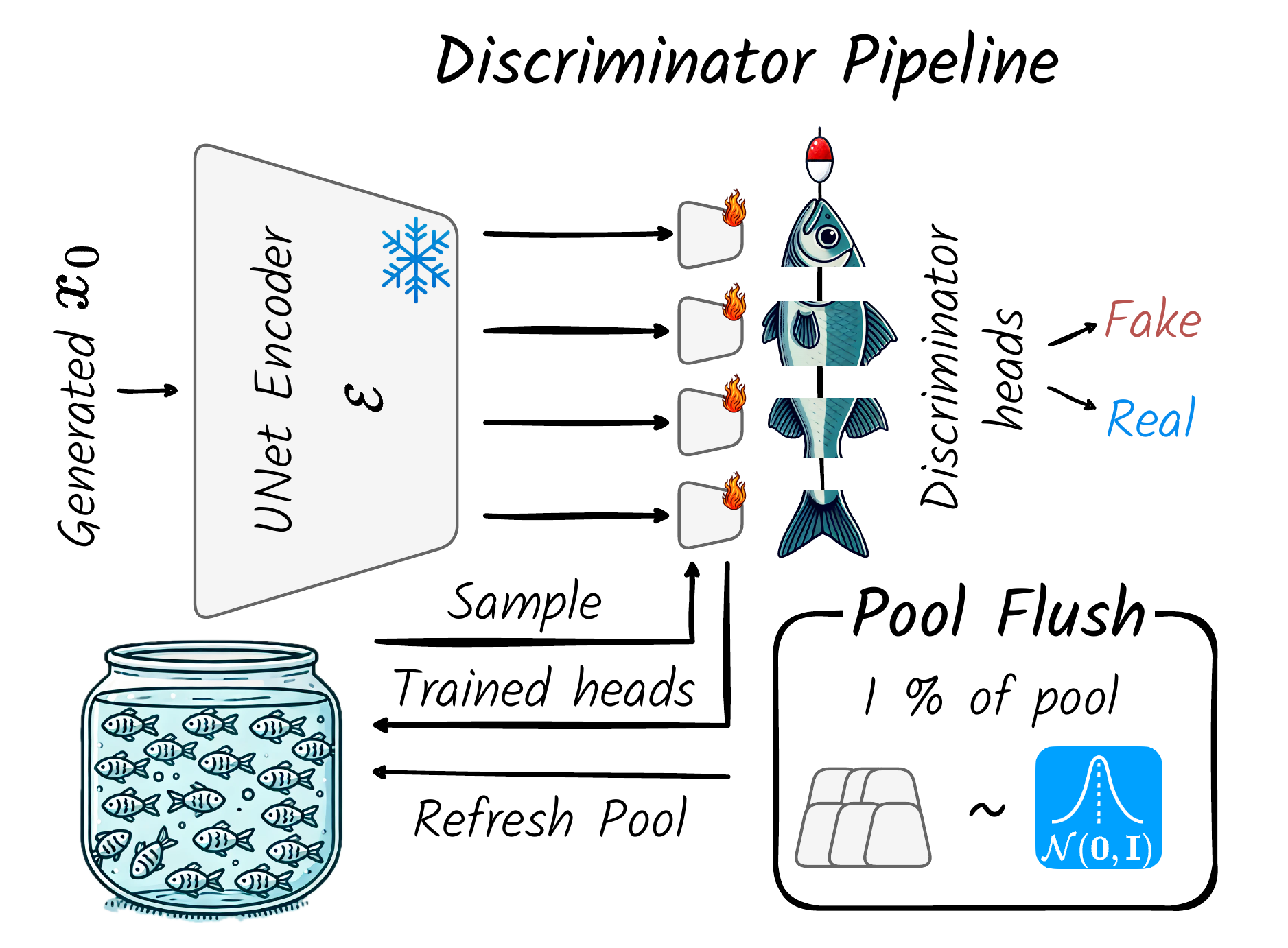}
    \vspace{-6mm}
    \caption{
    Our discriminator employs a frozen UNet backbone with a dynamic pool of discriminator heads. At each iteration, a subset of heads is sampled and trained, with 1\% of all heads randomly reinitialized to maintain diverse signals and prevent overfitting.
    }
   \vspace{-4mm}
    \label{fig:pool_head}
\end{figure}

Building on previous works~\cite{yin2024improved}, we utilize the teacher's ~\cite{rombach2022high} UNet encoder and mid-block as a frozen discriminator backbone $\mathcal{E}$ that extracts image features (see \cref{fig:pool_head}). This generally entails first noising inputs $x_0$ to pre-defined noise levels $t^*$ as $x_{t^*}$ and then using their denoising signals $\mathcal{E}(x_{t^*},t^*)$ as visual features. Different levels of the UNet encoder $\mathcal{E}$ provide feature representations at different levels, spanning from low-level details to high-level semantics. A lightweight trainable discriminator head is attached at each such level of the backbone $\mathcal{E}$ for the discriminator to perform \texttt{real}/\texttt{fake} classification.

As a core building block of our pipeline, we use a dynamic discriminator pool to source these discriminator heads. This discriminator pool $\mathcal{P}$ is a huge pool of constantly evolving discriminator heads that can be attached to $\mathcal{E}$ for our pipeline's multi-head discriminator. The lightweight design of these heads allows us to scale the pool without significant computational or memory overhead. For training the pool, we sample a subset of heads $\mathcal{D} \sim \mathcal{P}$ from the pool at every training iteration, computing the adversarial loss $\mathcal{L}_\text{adv}$ with this subset. We backpropagate gradients from $\mathcal{L}_\text{adv}$ to optimize the sampled heads $\mathcal{D}$. After the update, we release the heads back into the pool to evolve the global knowledge of the pool dynamically. The stochasticity of this process through random sampling ensures varied feedback, preventing any single head from dominating the generator’s learning and reducing bias. This diversifies feedback and enhances stability \cite{choi2022mclgan, albuquerque2019multi} in GAN training.

To construct specialized discriminator heads we compartmentalize the pool $\mathcal{P}$ based on the noise level of the discriminator timestep $t^*$ as $\{\mathcal{P}_{t^*} \in \mathcal{P} \; \forall \;  t^* \}$. This helps us sample discriminator heads $\mathcal{D}_{t^*} \sim \mathcal{P}_{t^*}$ that are specialized for a specific noise level at discriminator timestep $t^*$. Unlike prior approaches that treat timestep-dependent discriminators as augmentation or smoothing techniques ~\cite{wang2022diffusion, lin2024sdxl}, each head in our pool functions as an expert on its designated noise level, providing precise, nuanced critiques targeting specific image characteristics. We calculate the adversarial loss as: 
\begin{equation}
    \mathcal{L}^{G}_{\text{adv}} = -\mathbb{E}[ \Sigma_{\mathcal{H} \in \mathcal{D}_{t^*}} \mathcal{H}\left(\mathcal{E}(\hat{x}_{t^*}, t^*)\right) ]
\end{equation}
\begin{equation}
    \mathcal{L}^{D}_{\text{adv}} = \mathbb{E}[ \Sigma_{\mathcal{H} \in \mathcal{D}_{t^*}} \mathcal{H}\left(\mathcal{E}(\hat{x}_{t^*}, t^*)\right) - 
    \mathcal{H}\left(\mathcal{E}(x_{t^*}, t^*)\right) ]
\end{equation}
\noindent where the frozen UNet encoder $\mathcal{E}$ extracts features for sampled discriminator heads $\mathcal{D}_{t^*}$. Intermediate outputs from each trainable-head $\mathcal{H}$ are aggregated for \texttt{real}/\texttt{fake} discriminator predictions.

\vspace{-1mm}
\subsection{Discriminator Pool Refresh} 
\vspace{-1.5mm}
Early overfitting in GAN training limits the discriminator's feedback diversity, reducing the quality and variation of generated images ~\cite{sauer2023adversarial, lin2024sdxl, dao2024swiftbrushv2makeonestep}. To address this, we introduce a random re-initialization strategy for our dynamic discriminator pool: at each training iteration, we discard (flush) a random subset ($\sim$$1\%$) of discriminator heads, replacing (refreshing) them with re-initialized discriminators. Refreshing discriminator subsets helps maintain a balance between stable feedback from retained heads and variability from re-initialized ones to enhance generator performance.

\vspace{-1mm}
\subsection{Multi-Scale and Dual-Objective GAN Training}
\vspace{-1.5mm}
The generalization potential of diffusion models to multiple resolutions \cite{rombach2022high} allows us to further use the pre-trained UNet encoder for both global and local (patch) discrimination. For this, we divide the pool into local and global heads, training them with adversarial feedback -  to judge either the entire image, or fine-grained details respectively. This setup enables global-focused heads to assess structure and local-focused heads to capture textures, balancing macro and micro image details. 
We also introduce dual-objective GAN training which applies both conditional and unconditional adversarial loss.
We motivate this training following prior analysis \cite{lin2024sdxl} that confirms conditional generation to introduce ``Janus" artifacts while struggling to align images with text features. Janus artifacts present repeated patterns, such as faces or hands, within a local area. 
To reduce such artifacts that manifest more in single-step diffusions, we use local discriminator heads to perform conditional and unconditional discrimination.
Unconditional local heads provide feedback solely based on image coherence. This dual-objective approach prevents overfitting to specific prompt-driven features, reducing the likelihood of artifacts and delivering a balanced, generalized adversarial signal.

To summarize, we compartmentalize our pool of weights for each timestep $t^*$, where further boundaries are created for different training settings: (i) global images with conditional discrimination, (ii) local patches with conditional discrimination and (iii) local patches with unconditional discrimination. Each of these pools has the same number of discriminator heads.

\vspace{-1mm}
\subsection{Bottom-Up Multi-Step Refinement}
\vspace{-1.5mm}
Unlike previous step-reduction algorithms, we offer a quality v/s speed trade-off, where users can perform denoising on one-step or multiple steps (up to 4) to have higher-quality generated images with the same model weights. We support this by using a bottom-up refinement approach, where we optimize the network for one step, and iteratively refine for multiple steps one by one. This significantly differs from the more traditional top-down approaches that iteratively refine for $8, 4, 2$, and then $1$ step in that order. Using a bottom-up refinement approach allows users to use the same model for multiple steps, and obtain gradually improving results from $1$ to $4$ steps. 

\begin{figure}
\vspace{-0.3cm}
\begin{algorithm}[H]
\caption{Dynamic Adversarial Framework }
\begin{algorithmic} [1]
    \State \textbf{Input:} Teacher $G_{\psi}$, Student $G_{\theta}$, Pool $\mathcal{P}$, timesteps ${t^*_\text{all}}$ 

    \For{each timestep $t^* \in \{t*_\text{all}\}$ }
        \State \textbf{Initialize} $\mathcal{P}^{\text{global, uncond}}_{t^*}$, \{$\mathcal{P}^{\text{local, cond}}_{t^*}$, $\mathcal{P}^{\text{local, uncond}}_{t^*}\}$
    \EndFor

    \While{not converged}
        \State $\epsilon \sim \mathcal{N}(0,\mathbf{I})$
        \State \textbf{Sample Timestep:} $t^* \sim \{t^*_\text{all}\}$

        \State \textbf{Teacher output:} $x_0 \gets G_{\psi}(\epsilon)$
        \State \textbf{Student output:} $\hat{x}_0 \gets G_{\theta}(\epsilon)$
        \State $x_{t^*} \gets \sqrt{\bar{\alpha}_{t^*}} \cdot x_0 + \sqrt{1 - \bar{\alpha}_{t^*}} \cdot \epsilon  $
        \State $\hat{x}_{t^*} \gets \sqrt{\bar{\alpha}_{t^*}} \cdot \hat{x}_0 + \sqrt{1 - \bar{\alpha}_{t^*}} \cdot \epsilon$

        \For{compartment $\mathcal{P}^{\text{type}}_{t^*}$ in $\mathcal{P}_{t^*}$}
        \State $\mathcal{D}_{t^*} \sim \mathcal{P}^{\text{type}}_{t^*} $

        \State $\mathcal{L}^{D}_{\text{adv}} = \mathcal{D}_{t^*}(\mathcal{E}(\hat{x}_{t^*}, t^*)) - 
    \mathcal{D}_{t^*}(\mathcal{E}(x_{t^*}, t^*))$
    \State $\mathcal{L}^{G}_{\text{adv}} = -\mathcal{D}_{t^*}(\mathcal{E}(\hat{x}_{t^*}, t^*))$
        \State \textbf{Optimize}: $G_{\theta} - \alpha . \nabla\mathcal{L}_\text{adv}^G$
        \State \textbf{Optimize}: $\mathcal{P}^\text{type}_\text{optim} - \alpha . \nabla\mathcal{L}_\text{adv}^D$        
        \EndFor
        
        \State $\mathcal{P} \gets \{\mathcal{P}, \mathcal{P}_\text{optim}\}$
        \State $\mathcal{P}_\text{refresh} \sim \mathcal{N}(0,\mathbf{I})$
        
        \State $\mathcal{P} \gets \{\mathcal{P}, \mathcal{P}_\text{refresh}\}$
    \EndWhile

    \State \textbf{Return:} Trained student model $G_{\theta}$
\end{algorithmic}

\end{algorithm}
\vspace{-8mm}
\end{figure}

\begin{figure*}[ht!]
\begin{center}
   \includegraphics[width=.975\linewidth]{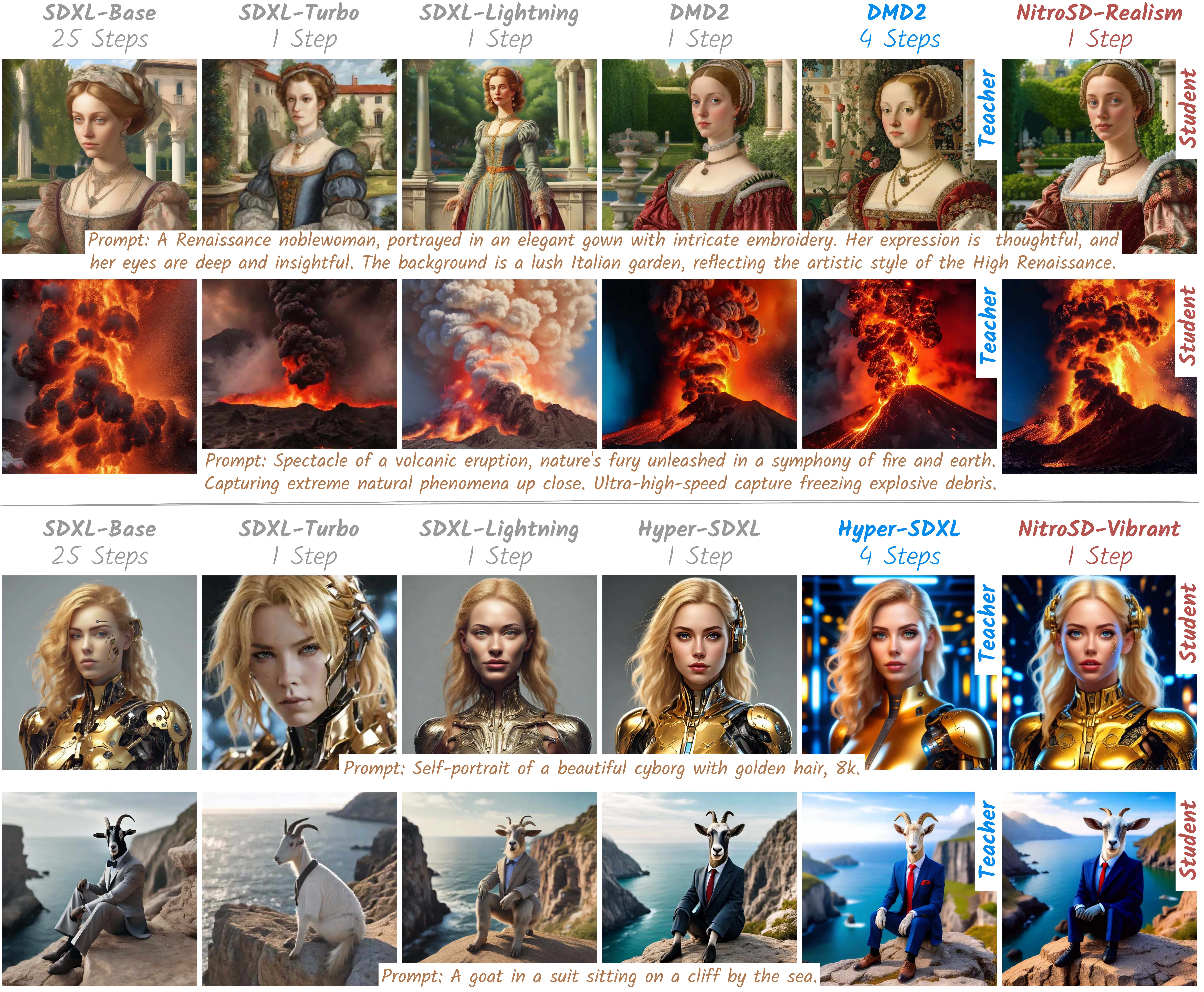}
\end{center}
   \vspace{-6mm}
   \caption{
   Visual comparison of our models (NitroSD-Realism and NitroSD-Vibrant) against multi-step SDXL~\cite{podell2023sdxl}, our teacher models (4-step DMD2~\cite{yin2024improved} and 8-step Hyper-SDXL~\cite{ren2024hyper}), and selected 1-step state-of-the-art baselines~\cite{sauer2023adversarial, lin2024sdxl}.
   }
   \vspace{-0.5mm}
\label{fig:qualitative_comparison}
\end{figure*}

\vspace{-2mm}
\section{Experiments}               
\label{sec:experiments}
\vspace{-1.5mm}

\begin{figure*}[t!]
\begin{center}
   \includegraphics[width=0.95\linewidth]{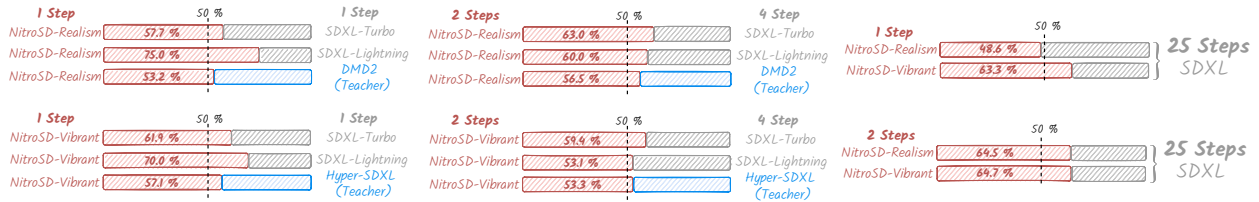}
\end{center}
   \vspace{-5.5mm}
   \caption{
   User preferences study with other baseline models.
   }
   \vspace{-4mm}
\label{fig:user_study}
\end{figure*}

\noindent\textbf{Implementation Details:}
Each discriminator head comprises $4 \times 4$ convolution layers with a stride of 2, group normalization~\cite{wu2018group}, and SiLU activation~\cite{ramachandran2017searching, hendrycks2023gaussian}. 10 heads work on 10 feature maps at different feature levels from a pretrained diffusion model’s frozen backbone. We employ specific discriminator timesteps $t^* \in \{10, 250, 500, 750\}$~\cite{lin2024sdxl}.

We use a pool of $480$ heads, using $160$ for each of the task types (global conditional / local conditional / local unconditional). We train using the  AdamW~\cite{loshchilov2019decoupledweightdecayregularization} optimizer with a batch size of 5 and gradient accumulation over 20 steps on a single NVIDIA A100 GPU.
Each iteration samples discriminator heads for real/fake classification from pool, with 1\% reinitialized (during pool refresh) to maintain dynamic feedback. To demonstrate generalization across teacher models, we train two networks with distinct visual goals: \textbf{NitroSD-Realism}, optimized for photorealism with the 4-step DMD2~\cite{yin2024improved} teacher; and \textbf{NitroSD-Vibrant}, for vivid colors with the 8-step Hyper-SDXL~\cite{ren2024hyper} teacher.

\begin{figure*}[htbp!]
\begin{center}
   \includegraphics[width=\linewidth]{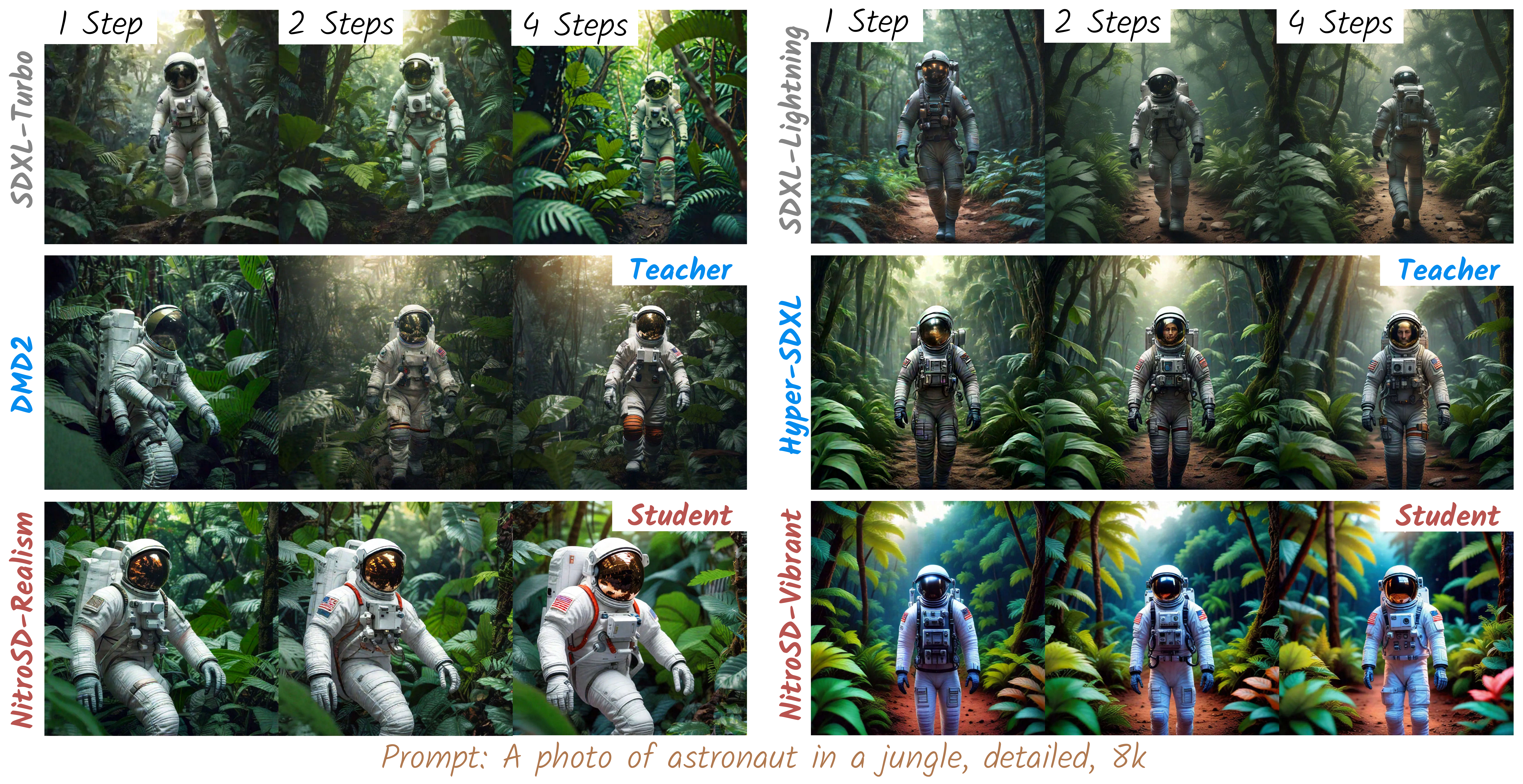}
\end{center}
   \vspace{-6.75mm}
   \caption{
    Visual comparison of our models (NitroSD-Realism and NitroSD-Vibrant) with other approaches across multiple steps, highlighting the clarity and improving quality of our method from 1-step to 4-step inference.
    }
   \vspace{-4mm}
\label{fig:multi_step}
\end{figure*}

\noindent\textbf{Data:} 
Following the hypothesis \cite{sauer2023adversarial} that synthetic images offer superior text alignment than real images, we train our models on synthetic samples only, generated by multi-step teacher models - without paired prompt-image data. Prompts are sourced from the Pick-a-Pic~\cite{kirstain2023pickapic} and LAION~\cite{schuhmann2022laion} datasets, totaling one million.

\noindent\textbf{Baseline Models and Evaluation Metrics:}
We compare our models to \textbf{DMD2}~\cite{yin2024improved}, \textbf{Hyper-SDXL}~\cite{ren2024hyper}, the \textbf{SDXL} base model~\cite{podell2023sdxl}, and additional timesteps distillation methods like iz\textbf{SDXL-Turbo}~\cite{sauer2023adversarial} and \textbf{SDXL-Lightning}~\cite{lin2024sdxl}.
\textit{DMD2}~\cite{yin2024improved} proposes distribution matching distillation using KL-divergence to address limitations in flow-guided distillation. \textit{Hyper-SDXL}~\cite{ren2024hyper} uses human feedback~\cite{xu2024imagereward, zhang2024unifl} to improve visual appeal of outputs.\textit{ SDXL-Turbo}~\cite{sauer2023adversarial} and \textit{SDXL-Lighting}~\cite{lin2024sdxl} 
introduce adversarial loss and timestep-dependent discriminator for low-step inference.

\vspace{-1mm}
\subsection{Qualitative Comparison}
\label{sec:qual_comparison}
\vspace{-1.5mm}

\Cref{fig:qualitative_comparison} provides a qualitative comparison of our models NitroSD-Realism and NitroSD-Vibrant against state-of-the-art diffusion models for one-step inference. Models \textit{SDXL-Turbo}~\cite{sauer2023adversarial} and \textit{SDXL-Lightning}~\cite{lin2024sdxl}, show limitations in visual fidelity. \textit{SDXL-Turbo} exhibits occasional text misalignment (e.g., 4th row), while \textit{SDXL-Lightning} often lacks sharpness in fine details.
In contrast, NitroSD-Realism and NitroSD-Vibrant exhibit greater clarity, richer textures, and fewer artifacts than all one-step benchmarks, including teacher models \textit{DMD2}~\cite{yin2024improved} and \textit{Hyper-SDXL}~\cite{ren2024hyper}. We also note that our models can pick up visual detail and texture fidelity of multi-step teachers, specifically \textit{Hyper-SDXL}’s 8-step and \textit{DMD2}’s 4-step models. NitroSD-Realism aligns closely with the photorealistic detail of \textit{DMD2}, reproducing fine-grained realism even in a single inference step. NitroSD-Vibrant captures the vibrant, saturated color characteristic of \textit{Hyper-SDXL}’s vivid style. This strong alignment in style and quality highlights the effectiveness of our proposed adversarial framework in distilling distinctive teacher attributes. Finally, we note in comparisons with \textit{SDXL}~\cite{podell2023sdxl}’s 25-step results that NitroSD achieves competitive detail and texture fidelity, effectively compressing \textit{SDXL}’s extensive process into a streamlined, one-step model without sacrificing visual quality.

\vspace{-1mm}
\subsection{User Study}
\vspace{-1.5mm}
We conduct a two-choice preference-based user study, illustrated in \Cref{fig:user_study}, where participants compare images generated by NitroSD-Realism and NitroSD-Vibrant against other one-step and multi-step methods. Our single-step results indicate that NitroSD-Vibrant consistently outperforms all models, including \textit{SDXL} with 25 steps, showcasing superior color vibrancy and richness. NitroSD-Realism also demonstrates strong performance, outperforming all one-step approaches.
We also evaluate our 2-step results against 4-step outputs from the same competitors observing a preference of our 2-step method against even 4-step baselines. This demonstrates NitroSD to achieve superior quality with fewer steps, and highlights the practical advantage of our framework for high-fidelity generation.

\vspace{-1mm}
\subsection{Quantitative Comparison}
\label{sec:quant_comparison}
\vspace{-1.5mm}

We conduct a quantitative evaluation on the COCO-5K validation dataset ~\cite{lin2014microsoft}, using several key metrics in \Cref{tab:quant}: CLIP score~\cite{radford2021learning} (ViT-B/32~\cite{dosovitskiy2021an}), which assesses prompt alignment by measuring the similarity between generated images and textual descriptions; Fréchet Inception Distance (FID)~\cite{heusel2017gans}, which evaluates image quality and diversity by comparing feature distributions of generated and real images; Aesthetic Score~\cite{aestheticscore}, which is trained on user preferences to quantify visual appeal; and ImageReward score~\cite{xu2024imagereward}, which reflects potential user preferences.

While FID and CLIP scores for our models are competitive,  NitroSD particularly excels in advanced metrics: Aesthetic Score and Image Reward. NitroSD-Realism outperforms its teacher DMD2~\cite{yin2024improved} both in Aesthetic Score and Image Reward, two metrics capturing image appeal and text alignment based on user preference. NitroSD-Vibrant also achieves one of the highest scores in these two metrics, reflecting its capability to produce visually engaging images that align with user preferences. 
These advanced metrics highlight NitroSD’s strengths in subjective quality, a critical factor in text-to-image generation. When paired with our user study findings, these results confirm that NitroSD effectively balances fast inference with high user satisfaction, offering a practical solution for applications that demand both efficiency and aesthetic appeal.

\begin{table}[!htbp]
\scriptsize
\centering
    \setlength\tabcolsep{4pt} 
    \begin{tabular}{lcccccc}
        \toprule
        Model & \makecell{Steps\\ ($\downarrow$)} & \makecell{CLIP \\($\uparrow$)}  &  \makecell{FID \\($\downarrow$)} & \makecell{Aesthetic\\ Score $(\uparrow)$}  & \makecell{Image \\ Reward($\uparrow$)}   \\
        \midrule
        SDXL-Base~\cite{podell2023sdxl}                               & 25  & 0.320 & 23.30 & 5.58 & 0.782 \\
        \midrule
        SDXL-Turbo~\cite{sauer2023adversarial}                                & 4  & 0.317 & 29.07 & 5.51 & 0.848 \\
        SDXL-Lightning~\cite{lin2024sdxl}                            & 4  & 0.312 & 28.95 & 5.75 & 0.749 \\
        Hyper-SDXL~\cite{ren2024hyper}                                & 4  & 0.314 & 34.49 & 5.87 & 1.091 \\
        DMD2~\cite{yin2024improved}                                      & 4  & 0.316 & 24.57 & 5.54 & 0.880 \\
        \textbf{NitroSD-Realism}                  & 4  & 0.313 & 29.09 & 5.60 & 0.945 \\
        \textbf{NitroSD-Vibrant}                  & 4  & 0.312 & 39.76 & 5.85 & 1.034 \\
        \midrule
        SDXL-Turbo~\cite{sauer2023adversarial}                                & 1  & 0.318 & 28.99 & 5.38 & 0.782 \\
        SDXL-Lightning~\cite{lin2024sdxl}                            & 1  & 0.313 & 29.23 & 5.65 & 0.557 \\
        Hyper-SDXL~\cite{ren2024hyper}                                & 1  & 0.317 & 36.77 & 6.00 & 1.169 \\
        DMD2~\cite{yin2024improved}                                      & 1  & 0.320 & 23.91 & 5.47 & 0.825 \\
        \textbf{NitroSD-Realism}                  & 1  & 0.320 & 25.61 & 5.56 & 0.856 \\
        \textbf{NitroSD-Vibrant}                  & 1  & 0.314 & 38.49 & 5.92 & 0.991 \\
        
        \bottomrule
    \end{tabular}
    \vspace{-.5mm}
    \caption{
    Quantitative Comparisons with State-of-the-Art Methods.
    }
    \vspace{-1mm}
\label{tab:quant}
\end{table}

\vspace{-1mm}
\subsection{Comparison on Multiple-Step Samples}
\vspace{-1.5mm}
We conduct comparisons on multi-step samples, as shown in \Cref{fig:multi_step}. Notably, models like \textit{SDXL-Lightning}~\cite{lin2024sdxl} and \textit{DMD2}~\cite{yin2024improved} lack a unified model for both one-step and multi-step inference, resulting in layout inconsistencies that limit users' ability to refine one-step outputs. \textit{Hyper-SDXL} sacrifices one-step performance to achieve a unified model. All approaches~\cite{sauer2023adversarial, lin2024sdxl, ren2024hyper, yin2024improved} aside from ours exhibit noticeable artifacts on complex scenes, particularly in areas with intricate textures, such as in lush vegetation or in the space suit of the astronaut in \cref{fig:multi_step}. When inference is extended to 4 steps, \textit{SDXL-Turbo} demonstrates significant degradation, showing its limitation at higher inference steps. In contrast, our models NitroSD-Realism and NitroSD-Vibrant exhibit high levels of image clarity and steadily improve fidelity from 1-step to 4-step.

\begin{figure}[!b]
\vspace{-3mm}
\begin{center}
   \includegraphics[width=1.\linewidth]{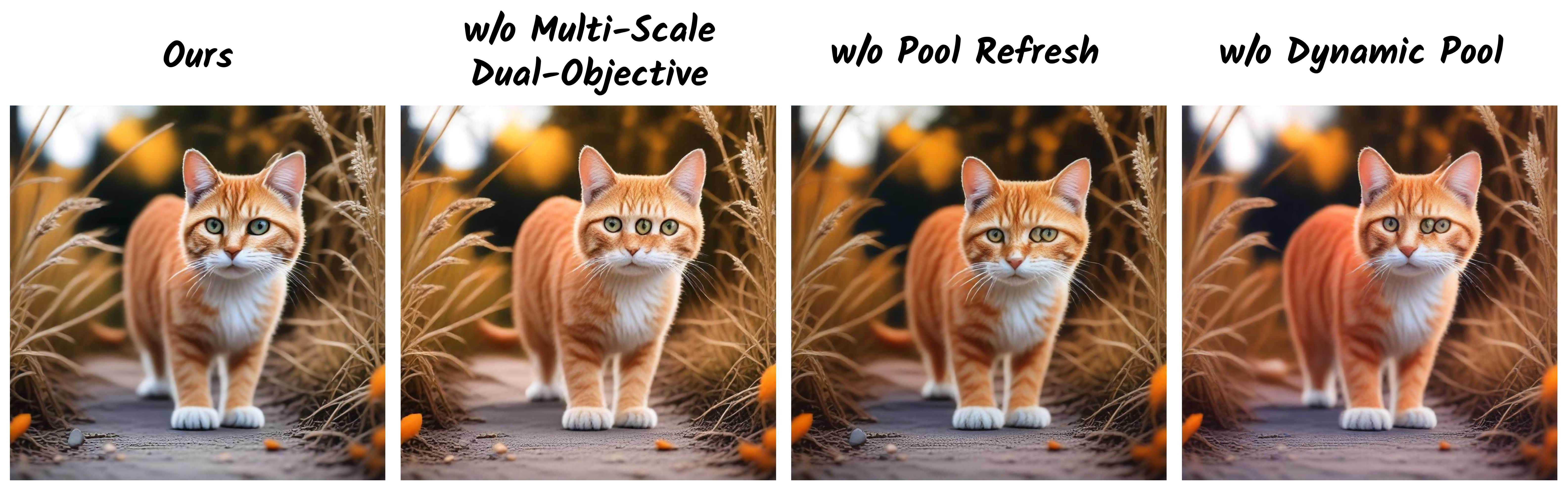}
\end{center}
   \vspace{-5mm}
   \caption{
    Qualitative study of ablative configurations
    }
\label{fig:ablation}
\end{figure}

\subsection{Ablation Study}
\label{sec:ablation}
\vspace{-1.5mm}

To assess the impact of each component in our Dynamic Adversarial Framework, we conduct an ablation study by removing specific elements, as shown in \Cref{fig:ablation}. We note that
(i) The absence of Multi-Scale Dual-Objective GAN Training reduces fine-grained details and introduces prominent triple-eyes Janus artifacts, highlighting the importance of balanced feedback.
(ii) Without Pool Refresh, artifacts persist and sharpness is lost, yielding poorer image quality. This suggests overfitting and lack of adaptiveness in the discriminator.
(iii) Removing Dynamic Discriminator Pool further reduces sharpness, indicating the pivotal role of the huge discriminator pool in our framework.

\vspace{-1mm}
\subsection{Extending to Diverse Teacher Models}
\begin{figure}
\begin{center}
   \includegraphics[width=1.\linewidth]{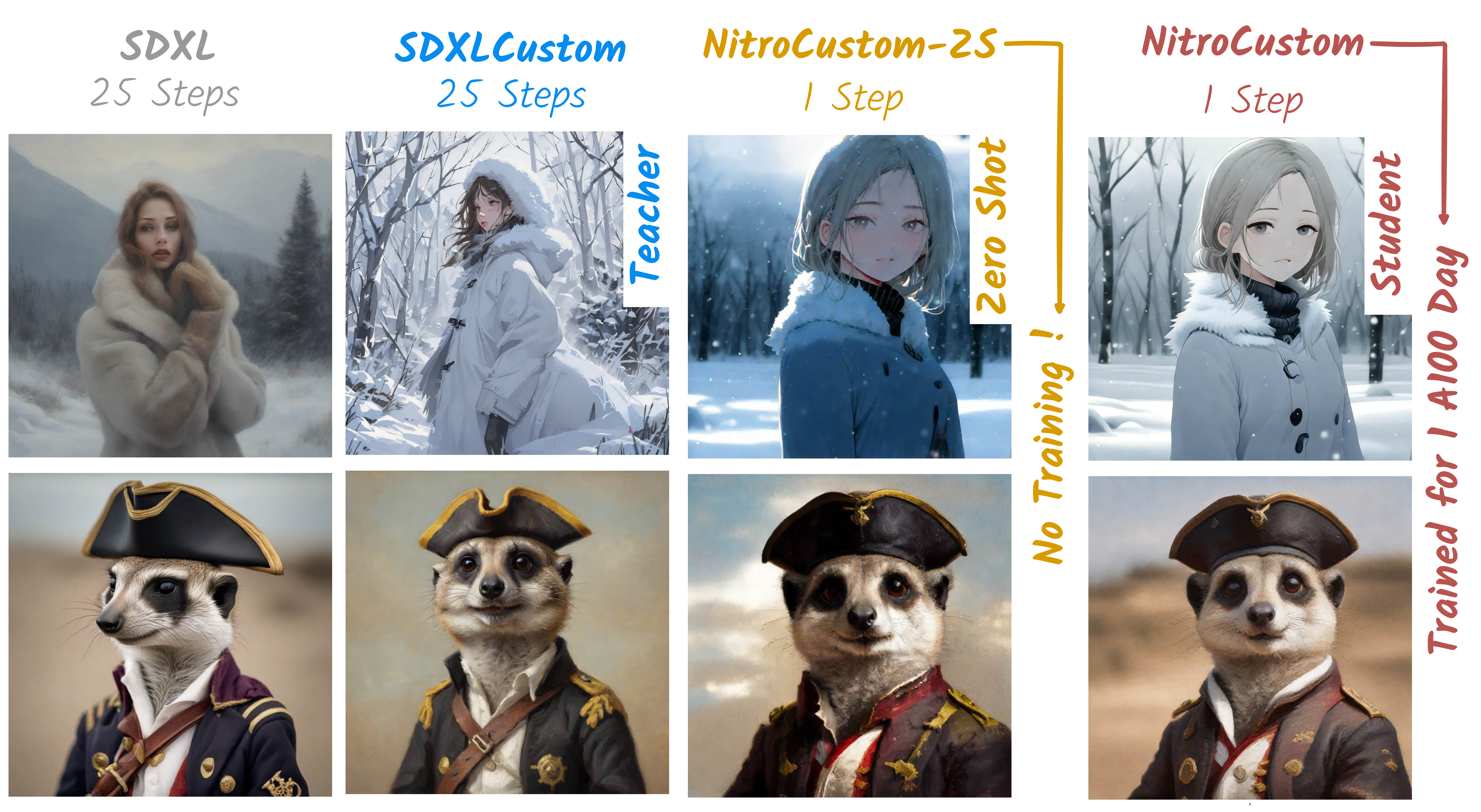}
\end{center}
   \vspace{-5mm}
   \caption{
    Results from applying NitroSD-Realism to anime~\cite{animaginexlv31} and oil painting~\cite{painterscheckpointv11} base models. Our model effectively adapts to different artistic styles.
    }
   \vspace{-4mm}
\label{fig:new_base}
\end{figure}

\vspace{-1.5mm}
Although NitroFusion is trained as a full model rather than as a LoRA~\cite{hu2021lora, db_lora}, it can adapt to other SDXL~\cite{podell2023sdxl} checkpoints through weight adjustment. This is achieved by applying the weight difference between NitroFusion and SDXL~\cite{podell2023sdxl} to a new custom model. \Cref{fig:new_base} illustrates results from adapting NitroSD-Realism to custom SDXL models having anime~\cite{animaginexlv31} and oil painting~\cite{painterscheckpointv11} styles from the CivitAI~\cite{civitai} community. Without additional training, NitroCustom-ZS (zero-shot) retains each style’s distinct characteristics using weight adjustments. NitroFusion’s independence from natural image data for training further allows easy adaptation to new styles (last column in \cref{fig:new_base})

\vspace{-1mm}
\section{Conclusion}
\label{sec:conclusion}
\vspace{-1mm}

In this paper, we propose a Dynamic Adversarial Framework for one-step diffusion distillation, using a huge pool of specialized discriminator heads to judge generation quality on multiple aspects - akin to a panel of art critics. We introduce a periodic refresh strategy for this pool, wherein a part of the pool is re-initialized to prevent discriminator overfitting and adversarial collapse. Finally, we train our entire setup with a multi-scale dual-objective strategy to focus on image detail at various scales (local v/s global) and balance prompt alignment with image coherence. Our model outperforms state-of-the-art low-step and one-step baselines in both qualitative and quantitative analysis. We perform extensive user studies and demonstrate that the majority of users prefer our one-step and two-step models, often even over 25-step high resolution diffusion pipelines.
{
    \small
    \bibliographystyle{ieeenat_fullname}
    \bibliography{main}
}

\clearpage
\setcounter{section}{0}
\renewcommand{\thesection}{\Alph{section}}
\newpage
\twocolumn[{
\centering
\Large
\textbf{\thetitle}\\
\vspace{0.5em}Supplementary Material \\
\vspace{1em}

\begin{center}
    \centering
    \captionsetup{type=figure}
    \includegraphics[width=1\linewidth]{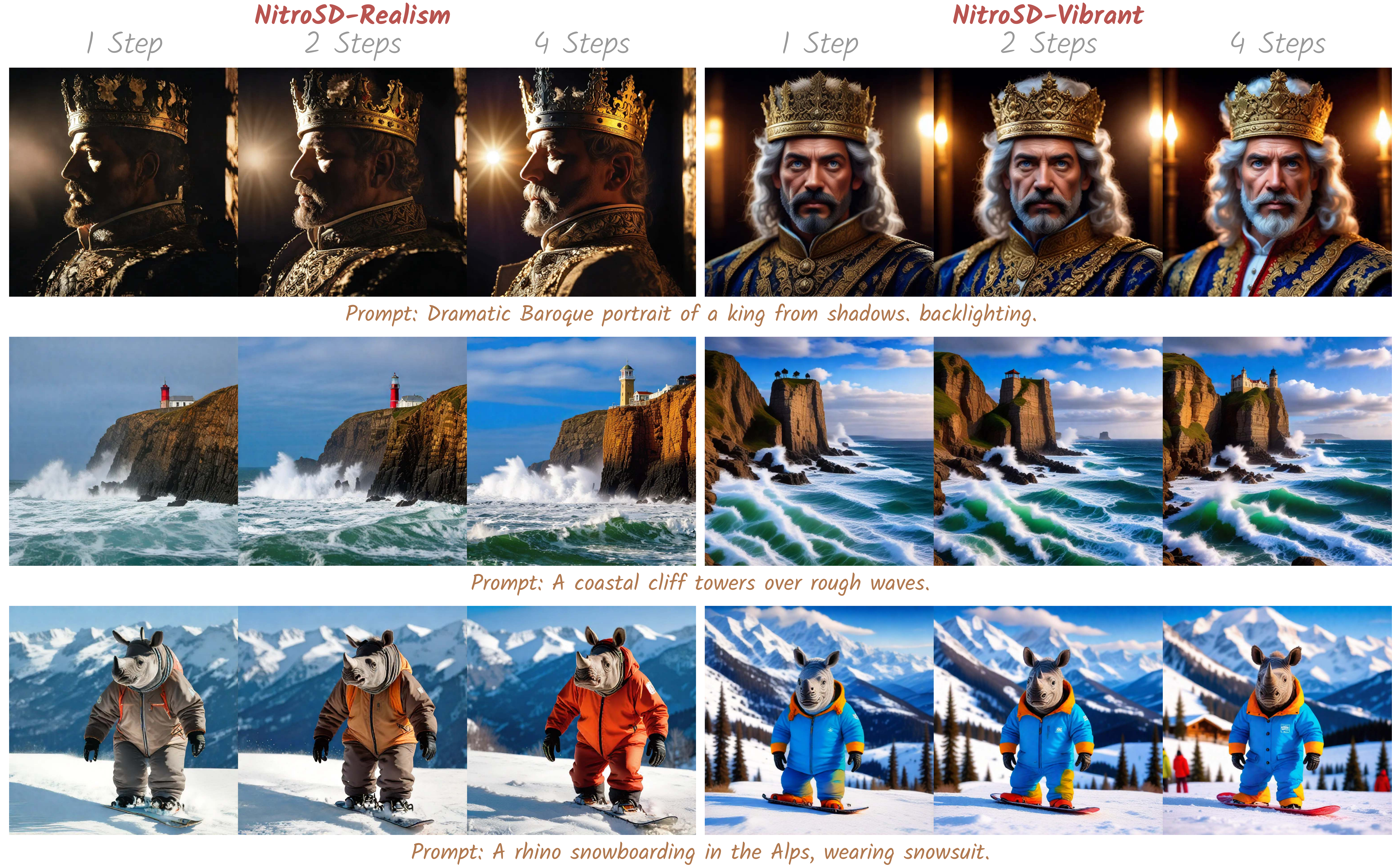}
    \vspace{-7mm}
    \captionof{figure}{
    1- to 4-step refinement process of our NitroSD-Realism and -Vibrant, illustrating the progressive enhancement of image quality and detail across steps.
    }
    \label{fig:multi_step_supp}
\end{center}
}] 

\section{Additional Implementation Details}
\label{sec:add_details}

\noindent\textbf{Timestep Shift:}  
Following prior works~\cite{chen2024pixart} and our base models, DMD2~\cite{yin2024improved} and Hyper-SD~\cite{ren2024hyper}, we adopt the timestep shift technique, shifting the original $T = 1000$ to $500$ and $250$. NitroSD-Realism and -Vibrant are trained on timesteps \{$250$, $188$, $125$, $63$\} and \{$500$, $375$, $250$, $125$\}, respectively, for multi-step generation. Both models were trained over approximately $20$ NVIDIA A$100$ days.

\noindent\textbf{User Study Details:}  
We evaluate user preferences using 128 prompts from the LADD~\cite{sauer2024fast} subset of PartiPrompts~\cite{yu2022scaling}, gathering 2,884 votes from 170 participants.

\begin{table}[!t]
\footnotesize
\centering
    \setlength\tabcolsep{4pt} 
    \begin{tabular}{lcccccc}
        \toprule
        Model & \makecell{CLIP \\($\uparrow$)}  &  \makecell{Patch Teacher\\FID ($\downarrow$)} & \makecell{Aesthetic\\ Score $(\uparrow)$}  & \makecell{Image \\ Reward($\uparrow$)}   \\
        \midrule
        Our Full                            & 0.315 & 18.70 & 5.87 & 1.020 \\
        w/o M-S D-O GAN                     & 0.316 & 18.99 & 5.83 & 1.035 \\
        w/o Pool Refresh                    & 0.316 & 18.78 & 5.98 & 1.054 \\
        w/o Dynamic Pool                    & 0.316 & 19.46 & 5.98 & 1.010 \\
        
        \bottomrule
    \end{tabular}
    \vspace{-2mm}
    \caption{
    Quantitative results of ablation study.
    }
    \vspace{-5mm}
\label{tab:quant_ablation_supp}
\end{table}

\section{Additional Ablation Study}
\label{sec:add_ablation}

The ablation study in \Cref{sec:ablation} employs the 8-step Hyper-SDXL~\cite{ren2024hyper} as the teacher, with 30 hours of training. \Cref{tab:quant_ablation_supp} presents the quantitative results. 

In particular, we introduce the Patch Teacher FID metric, which measures the FID score between $299 \times 299$ center-cropped patches from student and teacher samples~\cite{lin2024sdxl}, assessing how well high-resolution details are preserved. This metric serves as a critical index for evaluating the effectiveness of GAN training, as it emphasizes the generator's ability to represent fine-grained features and maintain fidelity to the teacher model.
\Cref{tab:quant_ablation_supp} shows that removing each component causes varying levels of degradation in Patch Teacher FID, highlighting the unique contributions of each to the overall performance of our Dynamic Adversarial framework.

\section{Discussion and Limitation}
\label{sec:limitation}

\noindent\textbf{Classifier-Free Guidance (CFG):}  
Like most few-step distillation methods~\cite{luo2023latent, ren2024hyper}, our framework does not support CFG~\cite{dhariwal2021diffusion, ho2022classifier}. While we achieve competitive results in one-step generation, incorporating CFG could enhance alignment with prompts, particularly for complex or ambiguous text. Future work could focus on integrating CFG into the adversarial framework to enhance controllability.

\noindent\textbf{Training with Natural Images:}  
Training on natural images offers the potential for improved quality by leveraging diverse, high-resolution data beyond teacher-generated samples. However, poorly aligned image-prompt pairs pose a significant risk of text-image misalignment, reducing adversarial training effectiveness. Future research will explore strategies for training with natural images while addressing image-prompt misalignment.

\noindent\textbf{Training Efficiency:}  
Our framework highlights the potential of adversarial training in one-step diffusion distillation, an area that remains underexplored. Future directions include optimizing adversarial strategies, such as more efficient adaptive learning schedules, to further boost training efficiency.

\section{Additional Qualitative Results}
\label{sec:add_qualitative}
\begin{figure*}[htbp!]
\begin{center}
   \includegraphics[width=\linewidth]{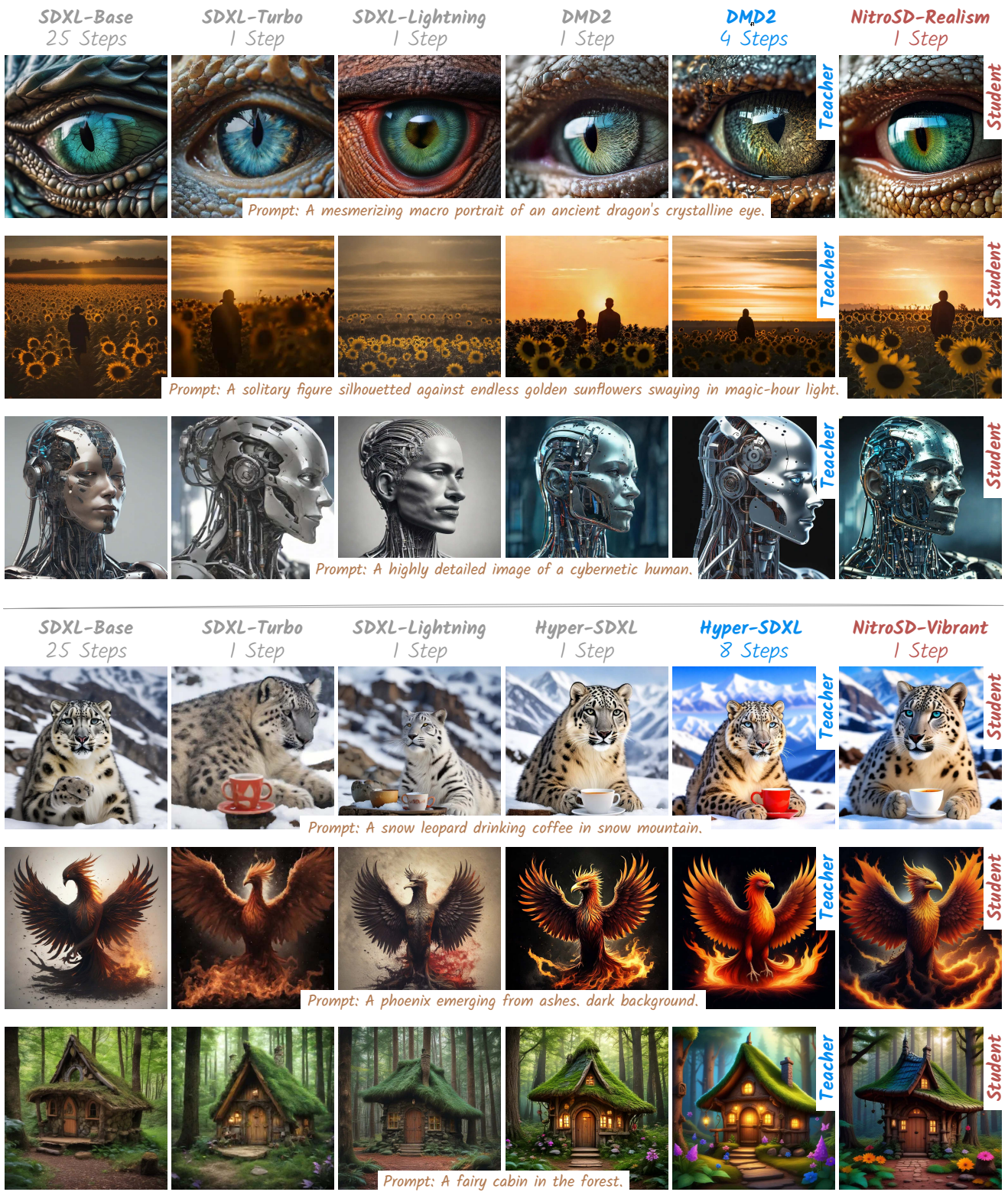}
\end{center}
   \vspace{-3mm}
   \caption{Additional visual comparison with state-of-the-art approaches.}
   \vspace{-2.5mm}
\label{fig:qualitative_comparison_supp}
\end{figure*}

\begin{figure*}[htbp!]
\begin{center}
   \includegraphics[width=0.9\linewidth]{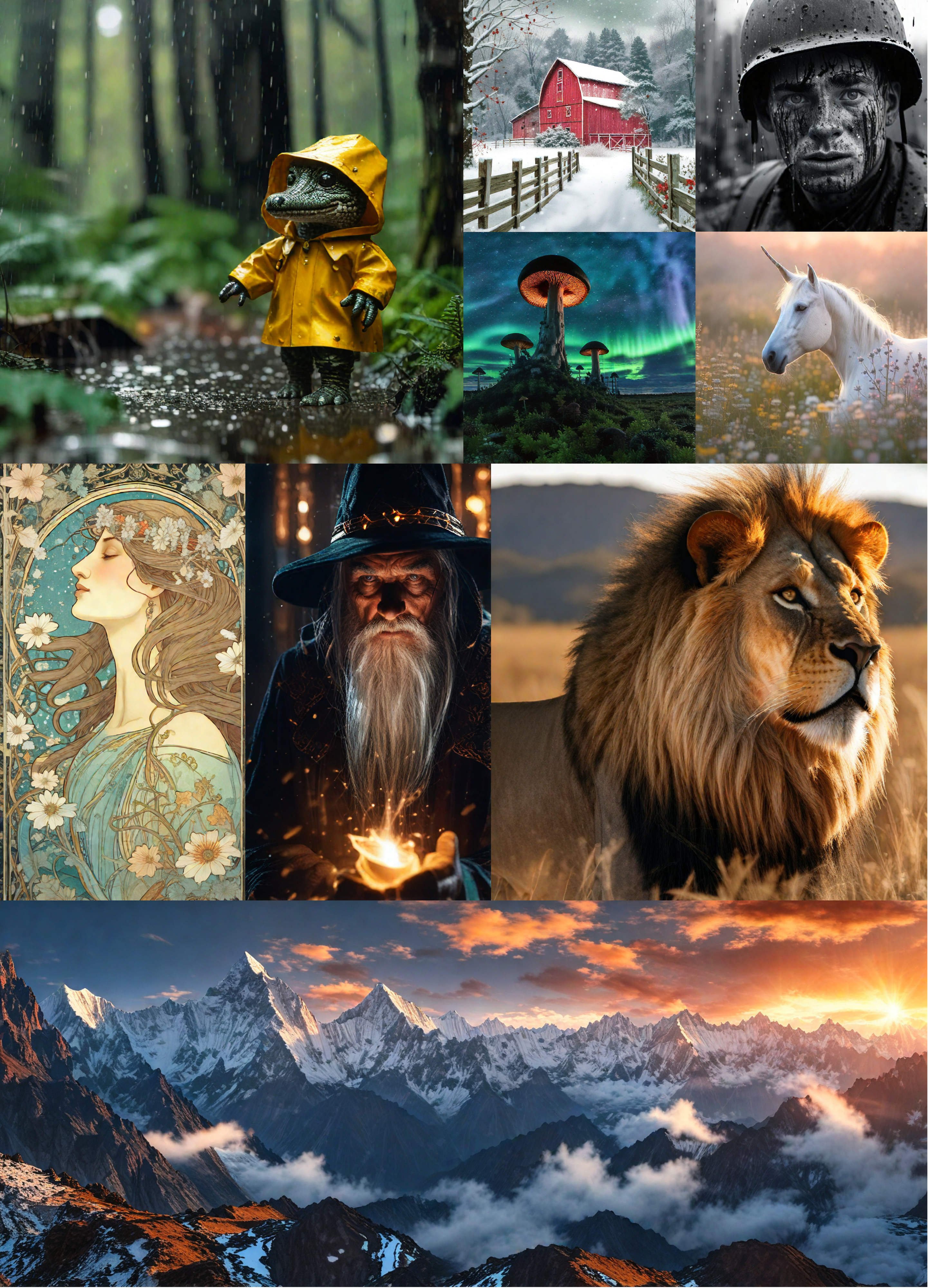}
\end{center}
   \vspace{-3mm}
   \caption{Additional single-step samples from NitroSD-Realism.}
   \vspace{-2.5mm}
\label{fig:qualitative_reslism_supp}
\end{figure*}

\begin{figure*}[htbp!]
\begin{center}
   \includegraphics[width=0.9\linewidth]{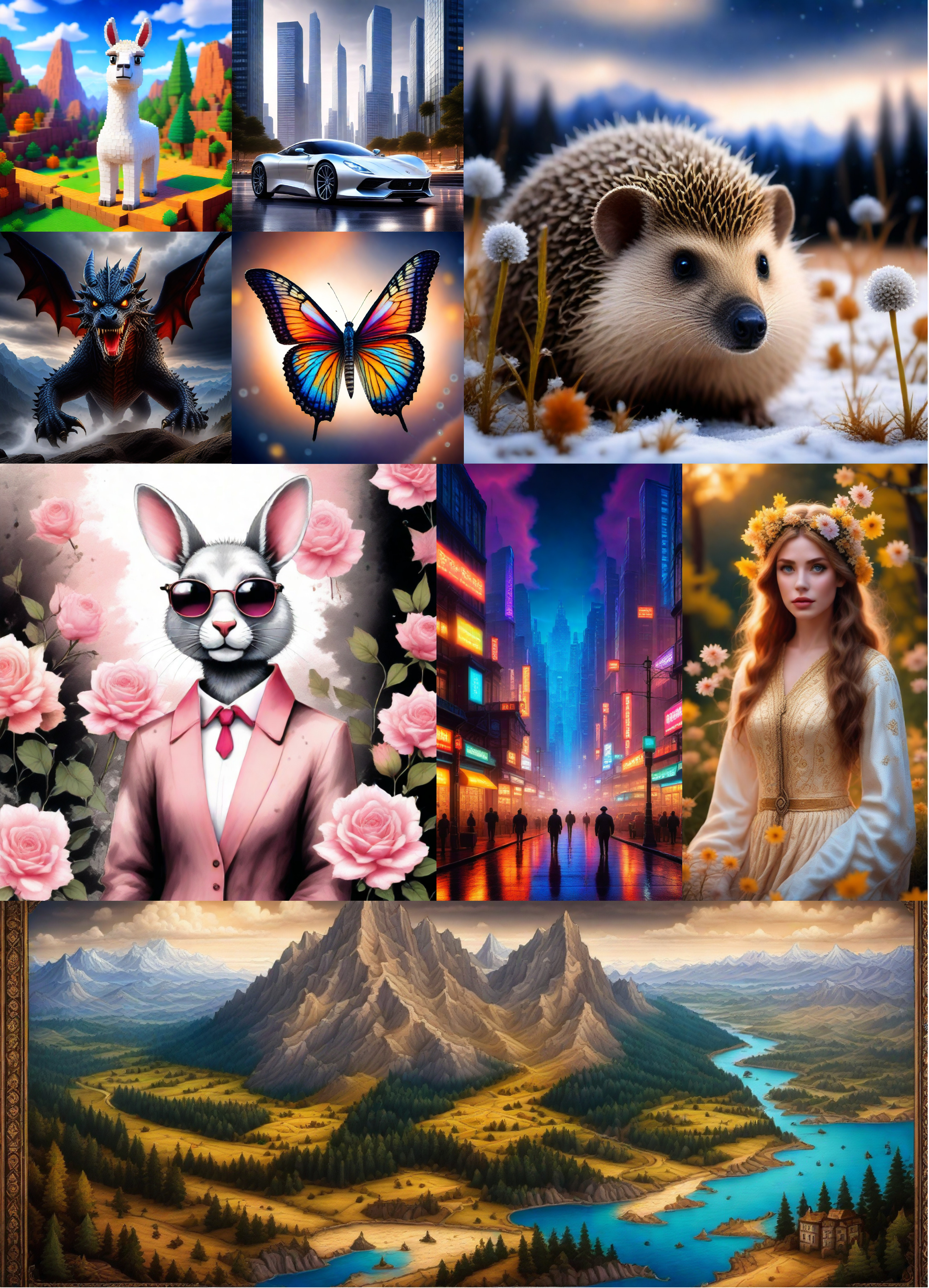}
\end{center}
   \vspace{-3mm}
   \caption{Additional single-step samples from NitroSD-Vibrant.}
   \vspace{-2.5mm}
\label{fig:qualitative_vibrant_supp}
\end{figure*}

We provide additional qualitative results in this section. \Cref{fig:multi_step_supp} showcases the 1- to 4-step refinement process of NitroSD, while \Cref{fig:qualitative_comparison_supp} presents further comparisons with baseline methods~\cite{podell2023sdxl, sauer2023adversarial, lin2024sdxl, ren2024hyper, yin2024improved}. Additionally, \Cref{fig:qualitative_reslism_supp} and \Cref{fig:qualitative_vibrant_supp} include more single-step samples generated by NitroSD-Realism and NitroSD-Vibrant, respectively.

\end{document}